\documentclass{article}

\usepackage{PRIMEarxiv}

\usepackage[utf8]{inputenc} 
\usepackage[T1]{fontenc}    
\usepackage{hyperref}       
\usepackage{url}            
\usepackage{booktabs}       
\usepackage{amsfonts}       
\usepackage{nicefrac}       
\usepackage{microtype}      
\usepackage{lipsum}
\usepackage{fancyhdr}       
\usepackage{graphicx}       
\graphicspath{{media/}}     

\usepackage{amsmath}
\usepackage{amssymb}
\usepackage{mathtools}
\usepackage{amsthm}
\usepackage{subdef}
\usepackage{float}
\usepackage{wrapfig}
\usepackage{microtype}

\title{Theoretical Analysis of Submodular Information Measures for Targeted Data Subset Selection} 

\author{
  Nathan Beck\thanks{Equal Contribution}\\
  University of Texas, Dallas \\
  \texttt{nathan.beck@utdallas.edu} \\
  \And 
  Truong Pham\footnotemark[1]\\
  University of Texas, Dallas \\
  \texttt{truong.pham@utdallas.edu} \\
  \And
  Rishabh Iyer\\
  University of Texas, Dallas \\
  \texttt{rishabh.iyer@utdallas.edu} \\
}

\begin{document}
\maketitle

\begin{abstract}
    With increasing volume of data being used across machine learning tasks, the capability to target specific subsets of data becomes more important. To aid in this capability, the recently proposed Submodular Mutual Information (SMI) has been effectively applied across numerous tasks in literature to perform targeted subset selection with the aid of a exemplar query set. However, all such works are deficient in providing theoretical guarantees for SMI in terms of its sensitivity to a subset's relevance and coverage of the targeted data. For the first time, we provide such guarantees by deriving similarity-based bounds on quantities related to relevance and coverage of the targeted data. With these bounds, we 
    show that the SMI functions, which have empirically shown success in multiple applications, are theoretically sound in achieving good query relevance and query coverage.
\end{abstract}

\section{Introduction}
\label{sec:intro}

As the volume of data used in various machine learning tasks increases in common usage, the ability to target and retrieve specific subsets of data becomes increasingly in-demand. One of the most common examples of targeted subset selection includes rare-class mining in data allocation phases, wherein subsets of a large corpus of data are chosen such that the subsets contain rare-class instances within some degree of expectation. Here, we focus our attention on the recently proposed information-theoretic concept of Submodular Mutual Information (SMI)~\cite{iyer2021submodular,iyer2021generalized}, which has been used to perform targeted subset selection and identification across a myriad of tasks with remarkable empirical efficacy. Briefly, SMI is defined as $I_F(A;\Qcal) = F(A) + F(\Qcal) - F(A\cup \Qcal)$, where $F$ is a submodular function (see Section~\ref{sec:smi}), and effectively generalizes the Shannon entropy mutual information function while measuring the information overlap of a subset $A$ and a query set $\Qcal$. The salient properties of SMI allow for a simple greedy approximation algorithm~\cite{nemhauser1978analysis} for maximizing $I_F(A;\Qcal)$ for a fixed $\Qcal$ under a cardinality constraint on $A$ in certain cases, which gives a simple yet highly versatile framework for performing targeted subset selection. Indeed, SMI and its maximization framework has been used in recent literature for rare-class active learning selection~\cite{kothawade2021similar, kothawade2022active, kothawade2022talisman,kothawade2022clinical,kothawade2022diagnose}, subset-guided domain adaptation~\cite{karanam2022orient}, weak supervision and retrieval~\cite{beck2024beyond}, semi-supervised meta learning~\cite{li2022platinum}, automatic speech recognition for rare accents~\cite{kothawade-etal-2023-ditto}, and multi-distributional active learning~\cite{beck2023streamline}.

\begin{wrapfigure}[48]{r}{0.45\textwidth}
    \centering
    \includegraphics[width=\linewidth]{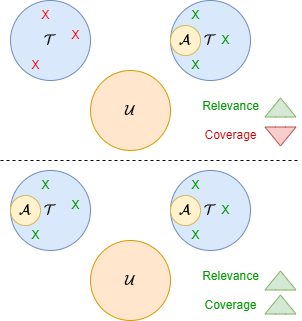}
    \caption{Illustration of the concepts of query relevance and query coverage. The top of the figure illustrates a scenario where $\Acal$ is relevant to the queries (X's) in the right cluster of $\Tcal$ but does not adequately cover the queries in the left cluster of $\Tcal$. The bottom of the figure illustrates a scenario where $\Acal$ covers all queries in both clusters.}
    \label{fig:metric_ex}
    \vspace{1ex}
    \includegraphics[width=\linewidth]{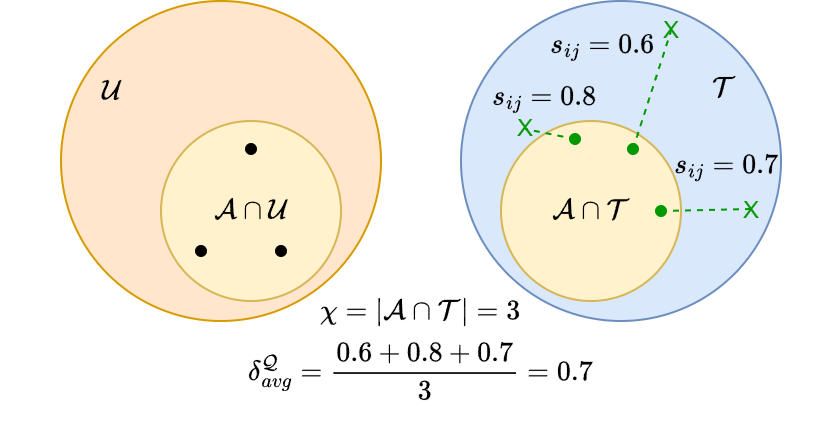}
    \caption{Illustration of the query relevance and query coverage metrics ($\chi$ and $\delta_{\text{avg}}^\Qcal$). Namely, we study query relevance through the lens of the number of targeted points selected (green points), which are those that match the cluster that the query points represent (green X's). We study query coverage by examining how well $A$ represents each point of $\Qcal$ on average (by using the most similar point to each query point).}
    \label{fig:delta_chi}
\end{wrapfigure}

Despite the empirical success of SMI's usage, no current work has provided any theoretical guarantee on the quality of the selected subset $A$ with respect to desirable quality metrics (Section~\ref{sec:smi}). Indeed, one may seek to ensure each instance of $A$ is close to instance in $\Qcal$ in a reasonable manner. This consideration leads to two metrics: \textit{Query relevance} and \textit{query coverage}. Query relevance measures the degree to which the instances of $A$ are in the same cluster(s) as the query set $\Qcal$. Figure~\ref{fig:metric_ex} depicts a scenario where $A$ exhibits high query relevance as it is within a targeted cluster (as shown by the presence of query instances within the same cluster). However, the top half of Figure~\ref{fig:metric_ex} also shows that not all query points have been \emph{covered}. Query coverage measures how well $A$ distributes over all the points in $\Qcal$. Indeed, the bottom half of Figure~\ref{fig:metric_ex} highlights a case with high query coverage as all query instances have instances in $A$ that are close by. While high values of these metrics are desirable, no explicit guarantees have been provided in past work much less any indication as to which assumptions would provide such guarantees. Hence, ascertaining the correct assumptions and developing guarantees on the quality of the selected subset in terms of these metrics would further strengthen the utility of SMI.

\newpage

To that end, we contribute for the first time guarantees on the selection quality of various similarity-based SMI instantiations in terms of query relevance and query coverage. For most of the SMI functions in~\cite{kothawade2022prism}, we derive an upper bound and a lower bound on their sensitivity to query relevance and query coverage; that is, our bounds directly relate the objective value of these functions to measurements of query relevance and query coverage (defined in Section~\ref{sec:smi}). The intuitions derived from these bounds directly matches past empirical observation, and we show these bounds to be tight across multiple simulated scenarios and hyperparameter configurations (Section~\ref{sec:relevance} and Section~\ref{sec:coverage}). This gives credence to the correctness of the bounds that we have derived and their usefulness in predicting the query relevance and query coverage behaviors of SMI functions. These theoretical underpinnings help solidify the application of SMI's query relevance and query coverage modeling capabilities, which have only been empirically observed in past work. 

\section{Submodular Mutual Information}
\label{sec:smi}

To motivate our discussion of Submodular Mutual Information~\cite{iyer2021submodular,iyer2021generalized}, we first start with the concept of submodular set functions. Given a ground set of instances $\Vcal$, a set function $F:2^\Vcal \rightarrow \mathbb{R}$ assigns a real-valued score to subsets of $\Vcal$. $F$ is said to be submodular~\cite{fujishige2005submodular,bilmes2022submodularity} if it exhibits the diminishing returns property: Given $A,B\subseteq \Vcal$ and $a \notin B$, $F(A \cup \{a\}) - F(A) \geq F(B \cup \{b\}) - F(B)$. Additionally, $F$ is said to be monotone if $F(A \cup \{a\}) \geq F(A)$. If $F$ is both monotone and submodular, then a simple greedy algorithm provides a $(1-\frac{1}{e})$-approximate solution for determining the cardinality-constrained subset $A$ that maximizes $F$~\cite{nemhauser1978analysis,iyer2019memoization}. Fortunately, as submodular functions are known to generalize other common information-theoretic functions such as Shannon entropy,~\cite{iyer2021submodular,iyer2021generalized} introduce the concept of submodular information measures that generalizes those based on Shannon entropy such as mutual information, conditional gain, and so forth. Of core interest is the Submodular Mutual Information (SMI) $I_F(A;Q)$, which is parameterized by an underlying submodular function $F$: $I_F(A;Q) = F(A) + F(Q) - F(A\cup Q)$. Intuitively, $I_F(A;Q)$ captures the amount of information overlap between a set $A\subseteq\Vcal$ and a set $Q\subseteq\Vcal$ (often called the \emph{query set}) based on the quantization of $F$. Importantly,~\cite{iyer2021submodular} show that $I_F(A;Q)$ is monotone submodular in $A$ for a fixed $Q$ under specific conditions for appropriate choices of $F$, meaning that the same greedy algorithm can be used to approximately solve for the cardinality-constrained $A$ which maximizes $I_F(A;Q)$. Lastly,~\cite{kothawade2022prism} extend the definition of SMI to allow for auxiliary query sets $Q\subseteq\Vcal'$ that need not be within the ground set via restricted submodularity (that is, submodularity only holds on a subset of $2^{\Vcal\cup\Vcal'})$.~\cite{kothawade2022prism} additionally provide many instantiations of SMI functions that have been subsequently used in many proceeding works~\cite{kothawade2021similar, kothawade2022diagnose, kothawade2022talisman, beck2023streamline, beck2024beyond, karanam2022orient}. We provide these instantiations in Table~\ref{tab:smis}.

We note that the SMI instantiations that have enjoyed widespread empirical use are based on similarity values between arbitrary elements $i$ and $j$ (as denoted by $s_{ij} \in [0,1]$ in Table~\ref{tab:smis}). Further, as the query set $Q$ used in SMI is an exemplar set of targeted data, we naturally consider a partitioning of the ground set into \emph{targeted} and \emph{untargeted} data as depicted in Figure~\ref{fig:metric_ex} and Figure~\ref{fig:delta_chi}. At a high level, we partition $\Vcal$ into a targeted class of instances $\Tcal$ and an untargeted class of instances $\Ucal$. This allows us to delineate scenarios by analyzing and bounding cross similarities between these partitions and the query set $\Qcal$. 

\begin{wraptable}[19]{r}{0.45\textwidth}
    \centering
    \caption{Instantiations of SMI functions using different base submodular functions~\cite{kothawade2022prism}. Each are based on pairwise similarities $s_{ij} \in [0,1]$ between an instance $i$ and $j$. Facility location is used to instantiate two variants: FLVMI and FLQMI. Graph cut (with parameter $\lambda$) is used to instantiate GCMI. A concave-over-modular function is used to instantiate COM (with concave function $\psi$).}
    \begin{tabular}{c|c}
         \hline
         \textbf{Name} & $I_F(A;Q)$ \\
         \hline 
         \hline
         FLVMI & $\sum\limits_{i\in\Vcal} \min \left( \max\limits_{j\in A} s_{ij}, \eta \max\limits_{j\in Q} s_{ij} \right)$ \\
         FLQMI & $\sum\limits_{i\in Q} \max\limits_{j\in A} s_{ij} + \eta \sum\limits_{i\in A} \max\limits_{j\in Q} s_{ij}$ \\
         GCMI & $2\lambda\sum\limits_{i\in A}\sum\limits_{j\in Q} s_{ij}$ \\
         COM & $\eta\sum\limits_{i\in A} \psi\left(\sum\limits_{j\in Q} s_{ij}\right) + \sum\limits_{i\in Q} \psi\left(\sum\limits_{j\in A} s_{ij}\right)$\\
         \hline
    \end{tabular}
    \label{tab:smis}
\end{wraptable}

Using this partitioning, we can now formally define how query relevance and query coverage can be measured for a given $A$ and $\Qcal$. Within our analysis, we measure query relevance through the number of targeted instances selected within $A$; specifically, we analyze $\chi = |A\cap\Tcal|$ as our query relevance measure. An example is given in Figure~\ref{fig:delta_chi}, where $A$ contains 3 targeted points (shown in green). We measure coverage of a set $S$ by averaging for each instance in $S$ the maximally similar element in $A$: $\delta_{\text{avg}}^S=\frac{1}{|S|} \sum_{i\in S}\max_{j\in A} s_{ij}$. If we let $S=\Qcal$, then this effectively measures query coverage. Figure~{2} also depicts how query coverage is calculated by averaging these maximally similar $A$ elements over each element in $\Qcal$.

Using these definitions, we are now equipped to provide a theoretical understanding of SMI's modeling capabilities of query relevance and query coverage. While the SMI instantiations of Table~\ref{tab:smis} have enjoyed widespread empirical use, little theoretical work has been put in place to discern whether the value of $I_F(A;Q)$ is attuned to the quality of the selected subset $A$ in terms of query relevance and query coverage. Broadly, we seek to answer the following: \textbf{1)} Under what scenarios is the value of $I_F(A;Q)$ sensitive to how relevant $A$ is to $Q$ using $\chi$? \textbf{2)} Under what scenarios is the value of $I_F(A;Q)$ sensitive to how well $Q$ is covered by $A$ using $\delta_{\text{avg}}^S$? In the following sections, we provide a number of theorems and takeaways that provide theoretical underpinnings of SMI's utility in modeling query relevance and query coverage.

\section{Takeaways and Main Results}
\label{sec:takeaways}

To root our theoretical analysis within the current literature, we provide our main findings within this section and relate them to past work; namely, we confirm a number of empirical findings from~\cite{kothawade2022prism} that lack theoretical backing by providing a number of theorems within Section~\ref{sec:relevance} and Section~\ref{sec:coverage}, where we expand in more detail upon our main findings. We additionally provide further analysis in Appendix~\ref{app:detail_relevance} and Appendix~\ref{app:detail_coverage}. We enumerate our main findings here:

\begin{enumerate}
    \item All of our theorems provide tight bounds on $\chi$ and $\delta_{\text{avg}}^S$, which measure query relevance and query coverage, respectively (Section~\ref{sec:smi}). The degree to which the SMI objective value is correlated with either metric depends on the variability between the lower and upper bounds that we have provided.
    \item Theorem~\ref{thm:flvmi_rel} and Figure~\ref{fig:relevance} shows that FLVMI exhibits the strongest correlation between its objective value and $\delta_{\text{avg}}^{\Tcal\setminus A}$, which is highly indicative of query coverage when $\Qcal$ represents $\Tcal$ well. Oppositely, FLVMI exhibits the weakest correlation between its objective value and $\chi$ as shown by Theorem~\ref{thm:flvmi_cov} and Figure~\ref{fig:coverage}. $\eta$ appears to have very little effect on these correlations. Hence, FLVMI is a good measure for query coverage but not query relevance, which corroborates the empirical findings of~\cite{kothawade2022prism}.
    \item Theorem~\ref{thm:flqmi_rel} and Theorem~\ref{thm:flqmi_cov} imply that FLQMI has moderate correlations between its objective value and both $\chi$ and $\delta_{\text{avg}}^{\Qcal}$. Figure~\ref{fig:relevance} and Figure~\ref{fig:coverage} confirm this finding, and subsequent analysis on $\eta$ shows that larger $\eta$ values induce stronger correlation between FLQMI and query relevance and weaker correlation between FLQMI and query coverage. Our analysis matches previous empirical observation with~\cite{kothawade2022prism}, which shows that FLQMI tends to lie between FLVMI and GCMI in its modeling capacity of query relevance and query coverage.
    \item Theorem~\ref{thm:gcmi_rel} and Theorem~\ref{thm:gcmi_cov} imply that GCMI's objective value is highly correlated with $\chi$ but poorly correlated with $\delta_{\text{avg}}^\Qcal$. Our analysis in Figure~\ref{fig:relevance} and Figure~\ref{fig:coverage} confirms this, which also supports previous empirical observation in~\cite{kothawade2022prism} that GCMI's objective value functions largely in the interest of query relevance. 
    \item Theorem~\ref{thm:com_rel} and Theorem~\ref{thm:com_cov} suggest that COM exhibits the same correlative behaviors as GCMI; however, increasing $\eta$ (which isn't present in GCMI) increases its correlation with $\chi$, allowing for some control over query relevance. Again, this supports previous empirical observation~\cite{kothawade2022prism}. 
\end{enumerate}

\newpage
\begin{wraptable}[30]{r}{0.5\textwidth}
    \centering
    \caption{Notations}
    \begin{tabular}{c|l}
        \hline
        \textbf{Term} & \textbf{Description}\\ 
        \hline
        \hline
        $\Tcal$ & Set of targeted instances\\
        $\Ucal$ & Set of untargeted instances\\
        $Q$ & Exemplar set of targeted instances\\
        $B$ & $|A|$\\
        $\psi$ & Strictly increasing concave function\\
        $\chi$ & $|A\cap\Tcal|$\\
        $\delta_{\text{avg}}^S$ & $\frac{1}{|S|} \sum\limits_{i\in S} \max\limits_{j\in A} s_{ij}$\\
        $\alpha_1,\beta_1$ & $\forall i\in\Ucal, \max\limits_{j\in\Qcal} s_{ij} \in [\alpha_1,\beta_1]$\\
        $\alpha_2,\beta_2$ & $\forall i\in\Tcal, \max\limits_{j\in\Qcal} s_{ij} \in [\alpha_2,\beta_2]$\\
        $\alpha_3$ & $\min\limits_{i\in \Tcal} \frac{1}{|\Qcal|}\sum\limits_{j\in \Qcal} s_{ij}$\\
        $\beta_3$ & $\frac{1}{|\Qcal|} \sum\limits_{i\in\Qcal} \max\limits_{j\in\Tcal} s_{ij}$\\
        $\alpha_4$ & $\frac{1}{|\Tcal\setminus A|} \sum\limits_{i\in\Tcal\setminus A} \min(\max\limits_{j\in A} s_{ij}, \eta\alpha_2)$\\
        $\beta_4$ & $\frac{1}{|\Tcal\setminus A|} \sum\limits_{i\in\Tcal\setminus A} \min(\max\limits_{j\in A} s_{ij}, \eta\beta_2)$\\
        $\gamma_1,\Delta_1$ & $\forall i\in\Ucal, \frac{1}{|\Qcal|} \sum\limits_{j\in\Qcal} s_{ij} \in [\gamma_1,\Delta_1]$\\
        $\gamma_2,\Delta_2$ & $\forall i\in\Tcal, \frac{1}{|\Qcal|} \sum\limits_{j\in\Qcal} s_{ij} \in [\gamma_2,\Delta_2]$\\
        $\gamma_3,\Delta_3$ & $\forall i\in\Qcal, \frac{1}{|A\cap\Ucal|} \sum\limits_{j\in A\cap\Ucal} s_{ij} \in [\gamma_3,\Delta_3]$\\
        $\gamma_4,\Delta_4$ & $\forall i\in\Qcal, \frac{1}{|A\cap\Tcal|} \sum\limits_{j\in A\cap\Tcal} s_{ij} \in [\gamma_4,\Delta_4]$\\
        $\Omega_\Ucal$ & $\min\limits_{i,j\in\Ucal} s_{ij}$\\
        $\Omega_{\Ucal\Tcal}$ & $\min\limits_{i\in\Ucal,j\in\Tcal} s_{ij}$\\
        $\Ocal$ & $\sum\limits_{i\in\Tcal\setminus A}(\max\limits_{j\in A} s_{ij} - \eta \max\limits_{j\in\Qcal} s_{ij}) \mathbf{1}_{\max\limits_{j\in A} s_{ij} > \eta \max\limits_{j\in\Qcal} s_{ij}}$\\
        \hline
    \end{tabular}
    \label{tab:variables}
\end{wraptable}

\section{SMI and Query Relevance}
\label{sec:relevance}

\begin{figure}[t]
    \centering
    \includegraphics[width=\textwidth]{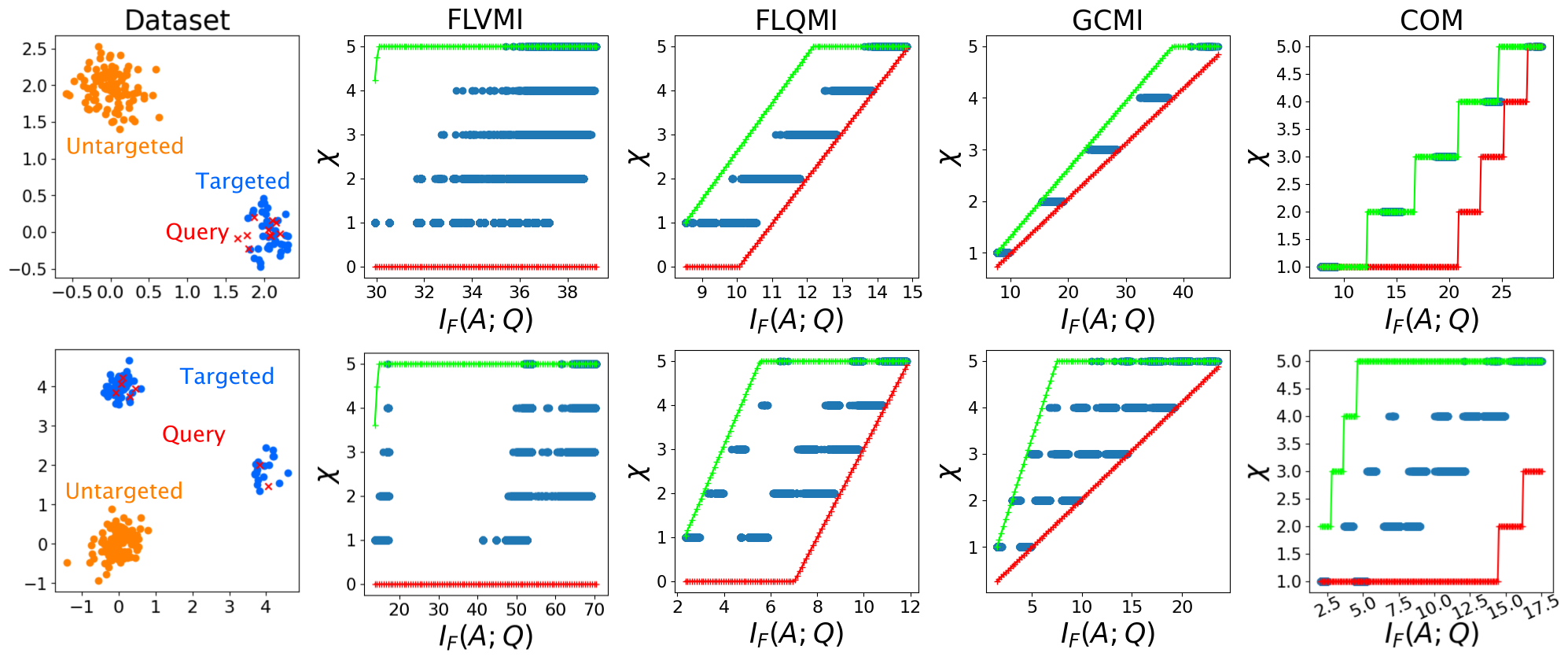}
    \caption{Behavior of the relevance bounds derived in Section~\ref{sec:relevance}. The synthetic datasets are generated by randomly sampling from different Gaussian distributions. Untargeted clusters are colored orange, targeted clusters are colored blue, and query instances are plotted as red X's. Using these dataset configurations, random subsets of cardinality $5$ are drawn with a uniform marginal distribution with respect to $\chi$, and the $I_F(A;Q)$ value is plotted against the $\chi$ (shown as blue points). The lower and upper bounds for each SMI function derived in Section~\ref{sec:relevance} are plotted in each subfigure and are clipped to be less than the budget (5) and greater than 0.}
    \label{fig:relevance}
\end{figure}

In this section, we present our theoretical analysis of the query relevance modeling of the SMI functions presented in Table~\ref{tab:smis}. Namely, we introduce bounds on $\chi$ as described in Section~\ref{sec:smi} utilizing similarity-based assumptions on the data. As each SMI function has varying characteristics, each bound assumes different similarity assumptions and refers to Table~\ref{tab:variables}. After presenting each theorem, we then discuss the implications of each bound and provide analysis as to the utility of each SMI function's query relevance modeling capabilities. We also depict empirically the relationship between $I_F(A;\Qcal)$ and $\chi$ for different functions and show our derived bounds in Figure~\ref{fig:relevance} to supplement our analysis. We also examine the effect of the hyperparameter $\eta$ on the correlation between $I_F(A;\Qcal)$ and $\chi$. Proofs of our theorems can be found in Appendix~\ref{app:relevance}, and additional analysis can be found in Appendix~\ref{app:detail_relevance}.

\begin{theorem}
    \label{thm:flvmi_rel}
    Let $A$ contain at least one targeted instance ($\chi \geq 1$). Using the notations of Table~\ref{tab:variables}, the Facility Location Mutual Information (FLVMI) enjoys the following bounds on $\chi$:
    \begin{align*}
        \chi &\geq \frac{I_F(A;\Qcal) - |\Ucal|\min(1,\eta\beta_1) - |\Tcal|\beta_4}{\min(1,\eta\beta_2) - \beta_4}\\
        \chi &\leq \frac{I_F(A;\Qcal) - |\Tcal|\alpha_4}{\min(1,\eta\alpha_2) - \alpha_4}
    \end{align*}
\end{theorem}

\begin{wrapfigure}[26]{r}{0.45\textwidth}
    \centering
    \includegraphics[width=\linewidth]{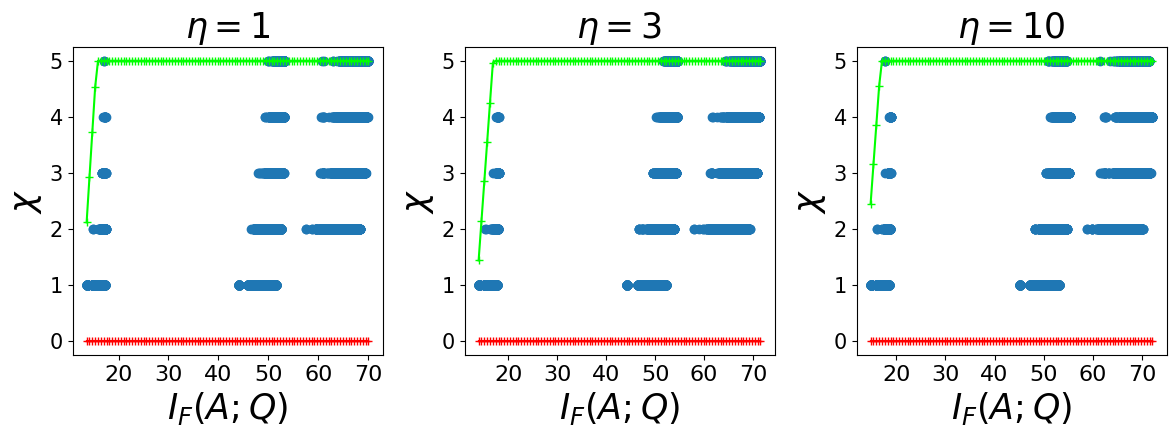}
    \caption{Effect of $\eta$ on the correlation between FLVMI's objective value and $\chi$ on the two-target dataset.}
    \label{fig:flvmi_rel_eta}
    \includegraphics[width=\linewidth]{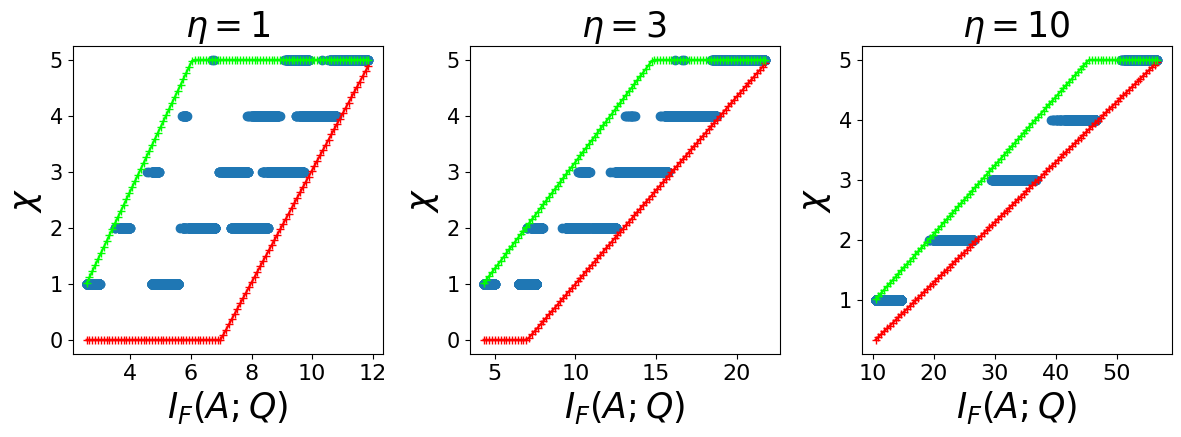}
    \caption{Effect of $\eta$ on the correlation between FLQMI's objective value and $\chi$ on the two-target dataset.}
    \label{fig:flqmi_rel_eta}
    \includegraphics[width=\linewidth]{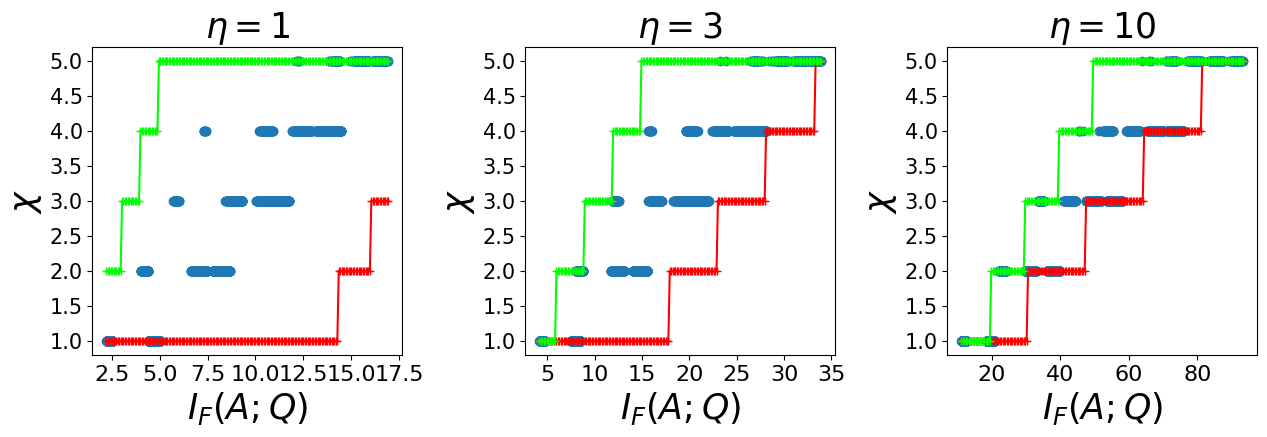}
    \caption{Effect of $\eta$ on the correlation between COM's objective value and $\chi$ on the two-target dataset.}
    \label{fig:com_rel_eta}
\end{wrapfigure}

\emph{The bounds in Theorem~\ref{thm:flvmi_rel} suggest that FLVMI does not correlate well to query relevance.} This is shown in Table~\ref{tab:relevance_r}, where it achieves the lowest Spearman rank correlation between $I_F(A;\Qcal)$ and $\chi$ of the studied SMI functions. The terms $|\Ucal|$ and $|\Tcal|$ in the numerator indicate a high sensitivity to the dissimilarities between the query and untargeted data ($\alpha_1$,$\beta_1$) and to the similarities between the query and targeted data ($\alpha_2$,$\beta_2$). From this analysis and the dataset in Figure~\ref{fig:relevance}, we can confirm that FLVMI can take high $I_F(A;\Qcal)$ values with low $\chi$ and vice versa, which is explained by the wide gap in its upper and lower bound. Lastly, we note that FLVMI's correlation with query relevance can be improved by increasing $\eta$ as shown in Figure~\ref{fig:flvmi_rel_eta} and in Table~\ref{tab:rel_eta}, but the effects are limited because $\eta$ is capped by the min terms in Theorem~\ref{thm:flvmi_rel}.

\begin{theorem}
    \label{thm:flqmi_rel}
    Let $A$ contain at least one targeted instance ($\chi \geq 1$). Using the notations of Table~\ref{tab:variables}, the Facility Location Variant Mutual Information (FLQMI) enjoys the following bounds on $\chi$ if $\alpha_1 < \alpha_2,\beta_1 < \beta_2$:
    \begin{align*}
        \chi &\geq \frac{I_F(A;\Qcal) - \eta B\beta_1 - |\Qcal|\beta_3}{\eta(\beta_2 - \beta_1)}\\
        \chi &\leq \frac{I_F(A;\Qcal) - \eta B\alpha_1 - |\Qcal|\alpha_3}{\eta(\alpha_2 - \alpha_1)}
    \end{align*}
\end{theorem}

In contrast to FLVMI, Theorem~\ref{thm:flqmi_rel} shows that FLQMI only depends on smaller multiplicative factors in the numerator of $|\Qcal|$ and B. \emph{This indicates a stronger correlation between FLQMI and $\chi$}, which is corroborated by Table~\ref{tab:relevance_r}. This intuition is also supported by Figure~\ref{fig:relevance}, where the bounds wrap tightly around the data samples. $\eta$ plays a more prominent role in increasing FLQMI's correlation strength with $\chi$ as depicted in Figure~\ref{fig:flqmi_rel_eta} and in Table~\ref{tab:rel_eta} as the $\eta$ terms first dominate the numerator and are subsequently stabilized by the denominator of the bounds in Theorem~\ref{thm:flqmi_cov}.

\begin{table}[h]
    \centering
    \caption{Spearman rank correlation coefficients between SMI objective value and relevance for each function given in Figure~\ref{fig:relevance}. Specifically, the samples are \textbf{1)} sorted by $I_F(A;\Qcal)$ value, \textbf{2)} ranked using the ordinal method to split ties, and \textbf{3)} are used to calculate Spearman rank correlation coefficients via the Pearson correlation coefficient formula.}
    \begin{tabular}{l|cccc}
        \hline
        \textbf{Dataset} & \textbf{FLVMI} & \textbf{FLQMI} & \textbf{GCMI} & \textbf{COM}\\
        \hline
        \hline
        One Target & 0.89241 & 0.99989 & 1.00000 & 1.00000\\
        Two Target & 0.85823 & 0.97504 & 0.97622 & 0.98970\\
        \hline
    \end{tabular}
    \label{tab:relevance_r}
\end{table}

\begin{theorem}
    \label{thm:gcmi_rel}
    Using the notations of Table~\ref{tab:variables}, the Graph Cut Mutual Information (GCMI) enjoys the following bounds on $\chi$ if $\gamma_1 < \gamma_2,\Delta_1 < \Delta_2$:
    \begin{align*}
        \chi &\geq \frac{\frac{1}{2\lambda |\Qcal|} I_F(A;\Qcal)-B\Delta_1}{\Delta_2 - \Delta_1}\\
        \chi &\leq \frac{\frac{1}{2\lambda |\Qcal|} I_F(A;\Qcal)-B\gamma_1}{\gamma_2 - \gamma_1}
    \end{align*}
\end{theorem}

\emph{Theorem~\ref{thm:gcmi_rel} suggests an even better correlation between relevance and objective value for GCMI than FLQMI}, which is confirmed in Table~\ref{tab:relevance_r} and in Figure~\ref{fig:relevance}. GCMI depends on a single $B$ term and thus experiences less variability between the upper and lower bounds. Consequently, this implies that $I_F(A;\Qcal)$ is more highly correlated with $\chi$. This lead to a very tight trend between query relevance and objective value, which is again supported by Figure~\ref{fig:relevance} and Table~\ref{tab:relevance_r}. Notably, GCMI does not have an $\eta$ hyperparameter and functions mainly in the interest of modeling query relevance.
\begin{wraptable}[9]{r}{0.5\textwidth}
    \centering
    \caption{Spearman rank correlation coefficients between SMI objective value and $\chi$ for varying $\eta$ values across functions as depicted in Figure~\ref{fig:flvmi_rel_eta}, Figure~\ref{fig:flqmi_rel_eta}, and Figure~\ref{fig:com_rel_eta}.}
    \begin{tabular}{l|ccc}
         \hline
         $\eta$ & \textbf{FLVMI} & \textbf{FLQMI} & \textbf{COM} \\
         \hline
         \hline
         1 & 0.86730 & 0.97430  & 0.99141\\
         3 & 0.87335 & 0.99968  & 0.99752\\
         10 & 0.83166 & 1.00000 & 0.99870\\
         \hline
    \end{tabular}
    \label{tab:rel_eta}
\end{wraptable}

\begin{theorem}
    \label{thm:com_rel}
    There exist strictly monotonic increasing functions $f_l,f_h$ such that $f_l(\chi) \leq I_F(A;\Qcal) \leq f_h(\chi)$ for Concave over Modular if $\gamma_2 > \gamma_1,\gamma_4>\gamma_3$ and $\Delta_2 > \Delta_1,\Delta_4>\Delta_3$.
\end{theorem}

Theorem~\ref{thm:com_rel} provides a different form of correlation guarantee than the previous theorems of this section. Namely, the bounding functions of $\chi$ can be used to generate feasible values of $I_F(A;\Qcal)$ for a specific value of $chi$. Hence, for a specific $I_F(A;\Qcal)$ value, one can determine which $\chi$ are feasible by apply Theorem~\ref{thm:com_rel}. Importantly, the monotonicity of the bounding functions ensure that increases in $I_F(A;\Qcal)$ lead to increases in feasible $\chi$ values, establishing a correlation. \emph{Although Theorem~\ref{thm:com_rel} is weaker, COM takes a similar form to GCMI due to the form of the bounding functions (shown in Appendix~\ref{app:relevance}) and should exhibit a lot of the same behaviors as GCMI}, which are indicated in Figure~\ref{fig:relevance} and Table~\ref{tab:relevance_r}. We conclude by noting increases in $\eta$ also result in increases in correlation strength between $I_F(A;\Qcal)$ and $\chi$ as shown in Figure~\ref{fig:com_rel_eta} and Table~\ref{tab:rel_eta}.

\section{SMI and Query Coverage}
\label{sec:coverage}

\begin{figure}[t]
    \centering
    \includegraphics[width=\textwidth]{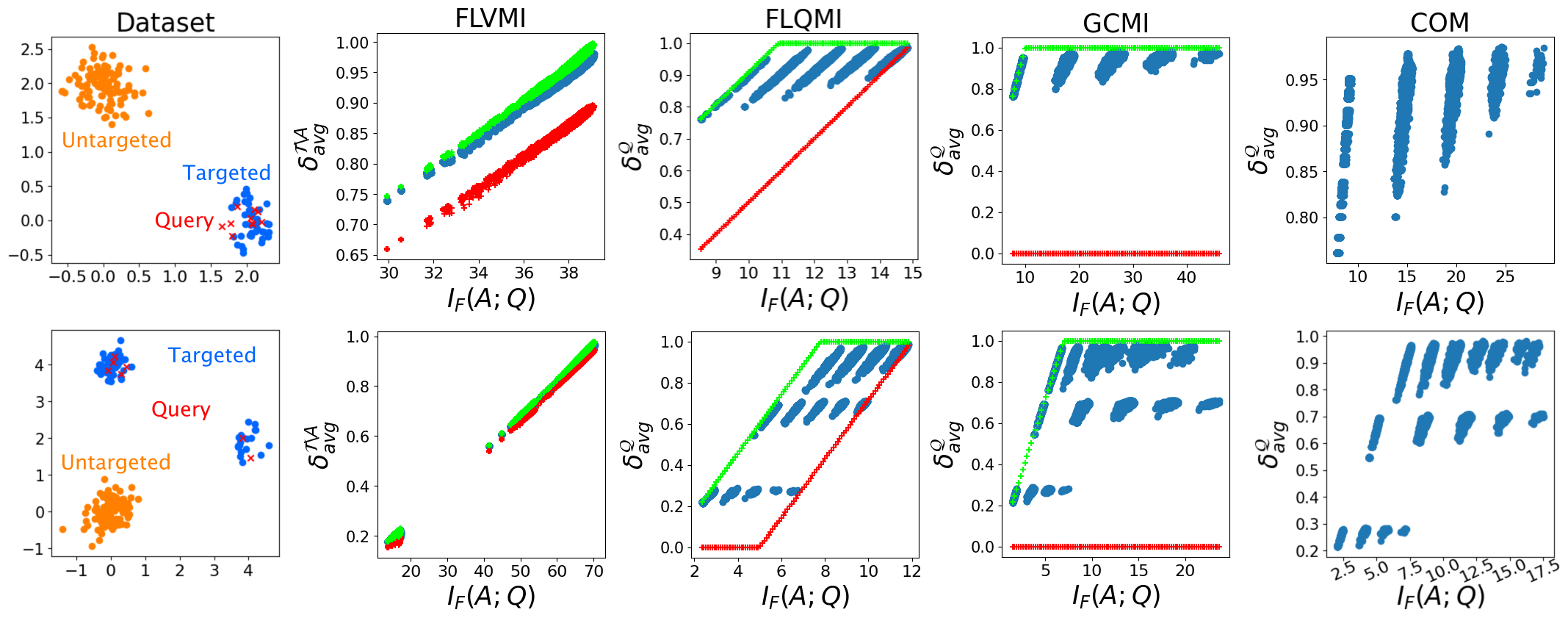}
    \caption{Behavior of the coverage bounds derived in Section~\ref{sec:coverage}. The synthetic datasets are generated by randomly sampling from different Gaussian distributions. Untargeted clusters are colored orange, targeted clusters are colored blue, and query instances are plotted as red X's. Using these dataset configurations, random subsets of cardinality $5$ are drawn with a uniform marginal distribution with respect to $\chi$, and the $I_F(A;Q)$ value is plotted against the $\delta_{\text{avg}}^S$ (shown as blue points). The lower and upper bounds for each SMI function derived in Section~\ref{sec:coverage} are plotted in each figure. Further, we apply a trivial clipping of the bounds as $\delta_{\text{avg}}^S$ must lie within 0 and 1.}
    \label{fig:coverage}
\end{figure}

In this section, we present our theoretical analysis of the query coverage modeling of the SMI functions presented in Table~\ref{tab:smis}. Namely, we introduce bounds on $\delta_{\text{avg}}^S$ for $S\in\{\Tcal,\Qcal\}$ as described in Section~\ref{sec:smi} utilizing similarity-based assumptions on the data. As before, each SMI function has varying characteristics, so each bound assumes different similarity assumptions and refers to Table~\ref{tab:variables}. We present each theorem, discuss the implications of each bound, and provide analysis as to the utility of each SMI function's query coverage modeling capabilities. We also depict empirically the relationship between $I_F(A;\Qcal)$ and $\delta_{\text{avg}}^S$ for different functions and show our derived bounds in Figure~\ref{fig:coverage} to supplement our analysis. We also examine the effect of the hyperparameter $\eta$ on the correlation between $I_F(A;\Qcal)$ and $\delta_{\text{avg}}^S$. Proofs of our theorems can be found in Appendix~\ref{app:coverage}, and additional analysis can be found in Appendix~\ref{app:detail_coverage}.

\begin{theorem}
    \label{thm:flvmi_cov}
    Let $A$ contain at least one targeted instance and one untargeted instance ($1 \leq \chi < B$). Using the notations of Table~\ref{tab:variables}, the Facility Location Mutual Information (FLVMI) enjoys the following bounds on $\delta_{\text{avg}}^{\Tcal\setminus A}:$
    \begin{align*}
        \delta_{\text{avg}}^{\Tcal\setminus A} &\geq \frac{I_F(A;\Qcal) - |\Ucal|\min(\eta\beta_1,1)}{|\Tcal| - \chi}\\
        & + \frac{\Ocal - \chi\min(\eta\beta_2,1)}{|\Tcal| - \chi}\\
        \delta_{\text{avg}}^{\Tcal\setminus A} &\leq \frac{I_F(A;\Qcal) - (B-\chi)\min(\eta\alpha_1,1)}{|\Tcal| - \chi}\\ 
        &- \frac{(|\Ucal|-B-\chi)\min(\max(\Omega_\Ucal,\Omega_{\Ucal\Tcal}),\eta\alpha_1)}{|\Tcal| - \chi}\\
        &+ \frac{\Ocal - \chi\min(\eta\alpha_2,1)}{|\Tcal|-\chi}
    \end{align*}
\end{theorem}

\begin{wrapfigure}[31]{r}{0.5\textwidth}
    \centering
    \includegraphics[width=\linewidth]{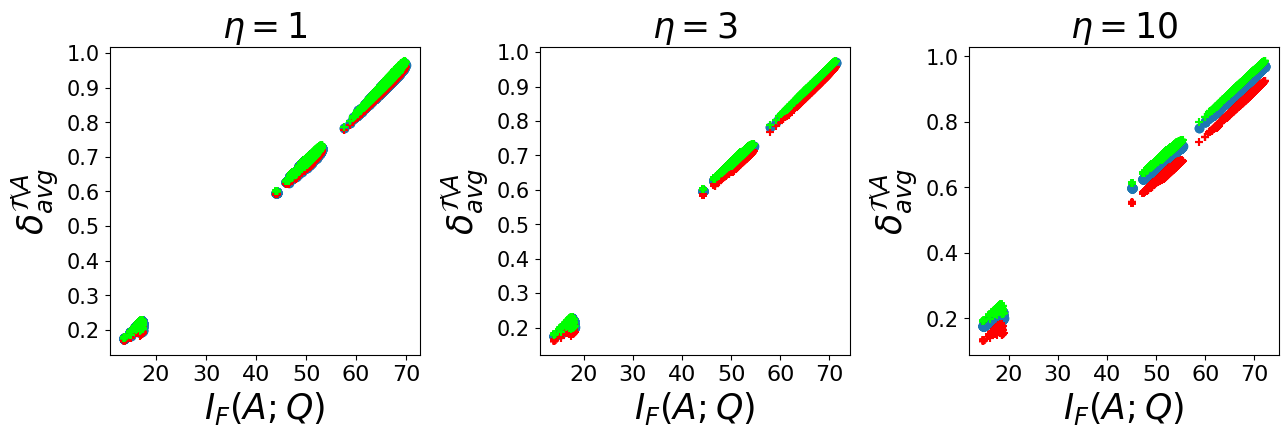}
    \caption{Effect of $\eta$ on the correlation between FLVMI's objective value and $\delta_{\text{avg}}^{\Tcal\setminus A}$ on the two-target dataset.}
    \label{fig:flvmi_cov_eta}
    \includegraphics[width=\linewidth]{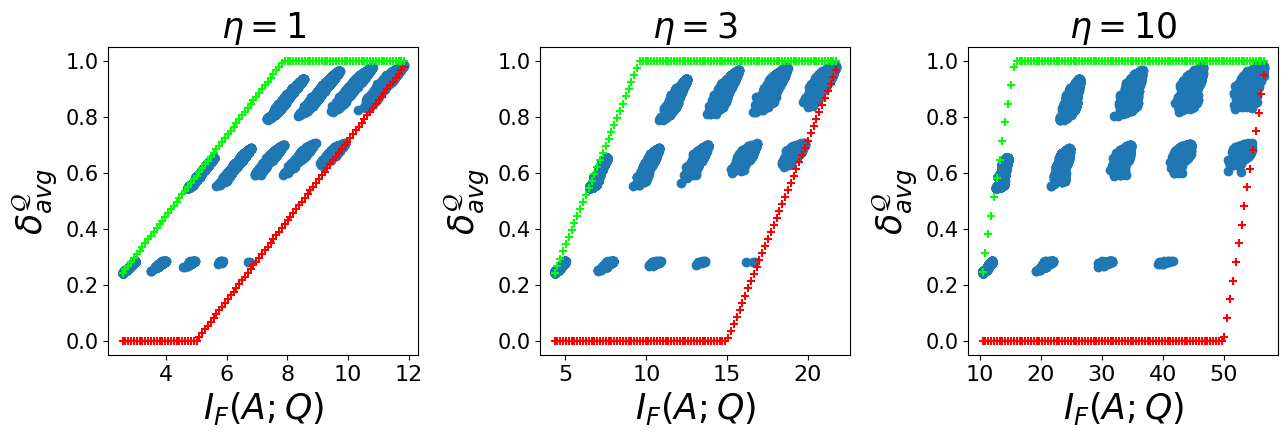}
    \caption{Effect of $\eta$ on the correlation between FLQMI's objective value and $\delta_{\text{avg}}^{\Qcal}$ on the two-target dataset.}
    \label{fig:flqmi_cov_eta}
    \includegraphics[width=\linewidth]{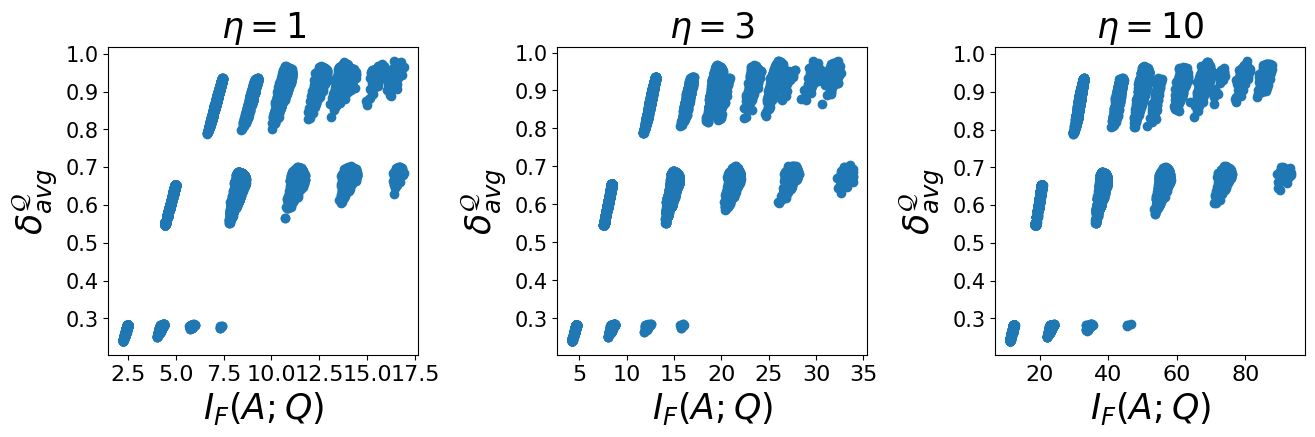}
    \caption{Effect of $\eta$ on the correlation between FLQMI's objective value and $\delta_{\text{avg}}^{\Qcal}$ on the two-target dataset.}
    \label{fig:com_cov_eta}
\end{wrapfigure}

\emph{While FLVMI's objective value is not well correlated with query relevance, it is more correlated with query coverage.} In Theorem~\ref{thm:flvmi_cov}, the larger $|\Ucal|$ terms enact upon $\alpha_1$ and $\beta_1$; hence, if the data is well-separated, these parameters should be small, so these terms should also be small. Additionally, the larger $\alpha_2$ and $\beta_2$ terms are mitigated by a smaller multiplicative factor of $\chi$. Combined with the normalization of $|\Tcal|$ in the denominator, Theorem~\ref{thm:flvmi_cov} suggests that the bounds are very stable. This is confirmed in Figure~\ref{fig:coverage} and in Table~\ref{tab:coverage_r}, where FLVMI exhibits the strongest correlation between $I_F(A;\Qcal)$ and $\chi$. Generally, high $\eta$ results in a slight decrease query coverage, but this effect is mitigated by the $\min$ function within the bounds. Such effect is visible in Figure~\ref{fig:flvmi_cov_eta} and in Table~\ref{tab:cov_eta}.

\begin{table}[t]
    \centering
    \caption{Spearman rank correlation coefficients between SMI objective value and coverage for each function given in Figure~\ref{fig:coverage}. Specifically, the samples are \textbf{1)} sorted by $I_F(A;\Qcal)$ value, \textbf{2)} ranked using the ordinal method to split ties, and \textbf{3)} are used to calculate Spearman rank correlation coefficients via the Pearson correlation coefficient formula.}
    \begin{tabular}{l|cccc}
        \hline
        \textbf{Dataset} & \textbf{FLVMI} & \textbf{FLQMI} & \textbf{GCMI} & \textbf{COM}\\
        \hline
        \hline
        One Target & 0.98986 & 0.82583 & 0.81854 & 0.81478\\
        Two Target & 0.99410 & 0.90883 & 0.58316 & 0.63301\\
        \hline
    \end{tabular}
    \label{tab:coverage_r}
\end{table}

\begin{theorem}
    \label{thm:flqmi_cov}
    Let $A$ contain at least one targeted instance ($\chi \geq 1$). Using the notations of Table~\ref{tab:variables}, the Facility Location Variant Mutual Information (FLQMI) enjoys the following bounds on $\delta_{\text{avg}}^{\Qcal}$:
    \begin{align*}
        \delta_{\text{avg}}^\Qcal &\geq \frac{I_F(A;\Qcal) - \eta(\chi\beta_2 + (B-\chi)\beta_1)}{|\Qcal|}\\
        \delta_{\text{avg}}^\Qcal &\leq \frac{I_F(A;\Qcal) - \eta(\chi\alpha_2 + (B-\chi)\alpha_1)}{|\Qcal|}
    \end{align*}
\end{theorem}

Theorem~\ref{thm:flqmi_cov} provides bounds that have the $\eta$ hyperparameter \emph{only} in the numerator. As a result, increasing $\eta$ will make the bounds on $\delta_{\text{avg}}^\Qcal$ more variable as there is no $\eta$ in the denominator to normalize. From previous analysis of Theorem~\ref{thm:flqmi_rel}, we know that increasing $\eta$ will make the bounds on $\chi$ \emph{less} variable. Hence, \emph{there is a clear trade-off in correlation strength between $I_F(A;\Qcal)$ and both $\chi$ and $\delta_{\text{avg}}^\Qcal$}. This is shown in Figure~\ref{fig:coverage}, which exhibits very similar plots as those in Figure~\ref{fig:relevance}. Additionally, the trade-off can be seen by comparing Figure~\ref{fig:flqmi_rel_eta}, Figure~\ref{fig:flqmi_cov_eta}, Table~\ref{tab:rel_eta}, and Table~\ref{tab:cov_eta}.

\begin{theorem}
    \label{thm:gcmi_cov}
    Using the notations of Table~\ref{tab:variables}, the Graph Cut Mutual Information (GCMI) enjoys the following bounds on $\delta_{\text{avg}}^{\Qcal}$:
    \begin{align*}
        \delta_{\text{avg}}^\Qcal &\geq \frac{1}{2\lambda|\Qcal|} I_F(A;\Qcal) - B\Delta_1 - \chi(\Delta_2 - \Delta_1)\\
        \delta_{\text{avg}}^\Qcal &\leq \frac{1}{2\lambda|\Qcal|} I_F(A;\Qcal) - (B - 1)\gamma_1 + \gamma_2 - \chi(\gamma_2 - \gamma_1)
    \end{align*}
\end{theorem}

\begin{wraptable}{r}{0.5\textwidth}
    \centering
    \caption{Spearman rank correlation coefficients between SMI objective value and $\delta_{\text{avg}}^S$ for varying $\eta$ values across functions as depicted in Figure~\ref{fig:flvmi_cov_eta}, Figure~\ref{fig:flqmi_cov_eta}, and Figure~\ref{fig:com_cov_eta}.}
    \begin{tabular}{l|ccc}
         \hline
         $\eta$ & \textbf{FLVMI} & \textbf{FLQMI} & \textbf{COM} \\
         \hline
         \hline
         1 & 0.99419 & 0.93301 & 0.66295\\
         3 & 0.99797 & 0.85515  & 0.65305\\
         10 & 0.99235 & 0.84374 & 0.65255\\
         \hline
    \end{tabular}
    \label{tab:cov_eta}
\end{wraptable}

Unlike GCMI's strong correlation between $I_F(A;\Qcal)$ and $\chi$, \emph{Theorem~\ref{thm:gcmi_cov} does not indicate strong correlation between $I_F(A;\Qcal)$ and $\delta_{\text{avg}}^\Qcal$}. The bounds depend heavily on how well separated $\Tcal$ and $\Ucal$ are and how similar $\Qcal$ is to $\Tcal$ ($\gamma_1$, $\gamma_2$,$\Delta_1$, $\Delta_2$). Further, as there are no denominators mitigating the effect of these attentuation terms, the bounds can have a large variance, indicating weak correlation between $I_F(A;\Qcal)$ and $\delta_{\text{avg}}^\Qcal$. This can be confirmed in Figure~\ref{fig:coverage} and Table~\ref{tab:cov_eta}, where GCMI clearly exhibits the weakest correlation (the coverage does not necessarily reach the maximum even with the highest objective value). 

\begin{theorem}
    \label{thm:com_cov}
    Let $A$ contain at least one targeted instance ($\chi \geq 1$). Then, there exist strictly monotonic $f_l,f_h$ for $i\in\Qcal$ such that the following holds for fixed $\chi$:
    \begin{align*}
        \sum_{i\in\Qcal} f_l(\max_{j\in A} s_{ij}) \leq I_F(A;\Qcal) \leq \sum_{i\in\Qcal} f_h(\max_{j\in A} s_{ij})
    \end{align*}
\end{theorem}

The statement of Theorem~\ref{thm:com_cov} has a slightly weaker guarantee on the correlation between $I_F(A;\Qcal)$ and $\delta_{\text{avg}}^\Qcal$ than in Theorem~\ref{thm:com_rel} since the bounding functions do not directly take in $\delta_{\text{avg}}^\Qcal$. However,
for sufficiently large increases in $\delta_{\text{avg}}^\Qcal$, it is reasonable to expect that all max terms would increase, lending to some
implicit correlation between $I_F(A;\Qcal)$ and $\delta_{\text{avg}}^\Qcal$. Due to the nature of the bounding functions (as shown in the proof in Appendix~\ref{app:coverage}), \emph{COM tends to once again mimic GCMI's correlation strength with $\delta_{\text{avg}}^\Qcal$}. This is confirmed in Figure~\ref{fig:coverage} and Table~\ref{tab:cov_eta}.  Lastly, $\eta$ shows limited impact with COM's correlation strength between $I_F(A;\Qcal)$ and $\delta_{\text{avg}}^\Qcal$ as shown in Figure~\ref{fig:com_cov_eta} and Table~\ref{tab:cov_eta}. This can be explained by the weaker guarantee afforded by Theorem~\ref{thm:com_cov}.

\section{Discussion}
\label{sec:discussion}

While we have shown theoretical results on SMI instantiations that have been applied across numerous applications, there are a number of interesting research directions for future study. Namely, one other SMI function that has been used in past work has not been analyzed -- Log Determinant Mutual Information (LOGDETMI): $I_F(A;\Qcal) = \log\det(S_A) + \log\det(S_\Qcal) - \log\det(S_{A\cup\Qcal})$, where $S_A$ denotes the matrix of similarity values between elements of $A$. Past work conducted in~\cite{kothawade2022prism} has provided empirical analysis that suggests LOGDETMI behaves similarly to FLQMI; however, showing theoretical justification for such a result tends to be challenging due to the fact that strong determinantal bounds are required that are connected to block submatrix substitutions of the notations in Table~\ref{tab:variables}. Another promising direction is to provide theoretical analysis of the other class of information-theoretic functions developed in~\cite{iyer2021submodular} -- that is, to show the modeling capabilities of Submodular Conditional Gain (SCG) and Submodular Conditional Mutual Information (SCMI) from a theoretical angle. Indeed, SCG (denoted $H_F(A|\Pcal)$) measures how semantically dissimilar $A$ is to a private set $\Pcal$, so a notion of privacy irrelevance could be argued for such SCG functions. Similarly, a combination of both query relevance, query coverage, and privacy irrelevance could be studied for SCMI (denoted $I_F(A;Q|\Pcal)$). We leave such considerations as an exciting avenue of future study.

In this paper, we provide theoretical guarantees on the selection quality of various SMI instantiations with respect to query relevance and query coverage by making assumptions on the average- and max-case similarities between $\Ucal$, $\Tcal$, and $\Qcal$. Through our theoretical analysis, we confirm a number of useful properties observed in previous work through the correlation between the objective value of several SMI instantiations and $\chi$ and $\delta_{\text{avg}}^S$, which reflect query relevance and query coverage, respectively. In general, we show that FLVMI's objective value is most attuned to query coverage and not well attuned to query relevance, even when the $\eta$ hyperparameter takes multiple different values. GCMI's objective value exhibits the opposite behavior of being attuned to query relevance but not query coverage, which COM roughly exhibits as well (with sufficiently high $\eta$). FLQMI falls in between these two extremes, effectively showing moderate correlation strength between $I_F(A;\Qcal)$ and both $\chi$ and $\delta_{\text{avg}}^\Qcal$, which can be attenuated by the $\eta$ hyperparameter. Importantly, our theoretical guarantees corroborate past empirical studies, further strengthening the utility of SMI.

\bibliographystyle{unsrt}  
\bibliography{main}  

\newpage
\appendix
\onecolumn

\section{Proofs of Query Relevance}
\label{app:relevance}

Here, we present the proofs of Theorem~\ref{thm:flvmi_rel}, Theorem~\ref{thm:flqmi_rel}, Theorem~\ref{thm:gcmi_rel}, and Theorem~\ref{thm:com_rel} as discussed in Section~\ref{sec:relevance}. For clarity, we repeat the statement of each theorem and include their proofs here. Note that each theorem refers to Table~\ref{tab:variables}.

\begingroup
\def\thetheorem{\ref{thm:flvmi_rel}}
\begin{theorem}
    Let $A$ contain at least one targeted instance ($\chi \geq 1$). Using the notations of Table~\ref{tab:variables}, the Facility Location Mutual Information (FLVMI) enjoys the following bounds on $\chi$:
    \begin{gather*}
        \chi \geq \frac{I_F(A;\Qcal) - |\Ucal|\min(1,\eta\beta_1) - |\Tcal|\beta_4}{\min(1,\eta\beta_2) - \beta_4}\\
        \chi \leq \frac{I_F(A;\Qcal) - |\Tcal|\alpha_4}{\min(1,\eta\alpha_2) - \alpha_4}
    \end{gather*}
\end{theorem}
\addtocounter{theorem}{-1}
\endgroup
\begin{proof}
    We obtain the result by separating the sum of FLVMI and substituting the notations of Table~\ref{tab:variables}:
    \begin{gather*}
        I_F(A;\Qcal) = \sum\limits_{i\in\Ucal} \min\left( \max\limits_{j\in A} s_{ij}, \eta \max\limits_{j\in\Qcal} s_{ij} \right) + \sum\limits_{i\in\Tcal\cap A} \min\left( \max\limits_{j\in A} s_{ij}, \eta \max\limits_{j\in\Qcal} s_{ij} \right) + \sum\limits_{i\in T\setminus A} \min\left( \max\limits_{j\in A} s_{ij}, \eta \max\limits_{j\in\Qcal} s_{ij} \right)\\
        \leq |\Ucal|\min(1,\eta\beta_1) + \chi\min(1,\eta\beta_2) + (|\Tcal| - \chi)\beta_4\\
        \implies\\
        \chi \geq \frac{I_F(A;\Qcal) - |\Ucal|\min(1,\eta\beta_1) - |\Tcal|\beta_4}{\min(1,\eta\beta_2) - \beta_4}\\
        I_F(A;\Qcal) \geq \chi\min(1,\eta\alpha_2) + (|\Tcal|-\chi)\alpha_4\\
        \implies\\
        \chi \leq \frac{I_F(A;\Qcal) - |\Tcal|\alpha_4}{\min(1,\eta\alpha_2) - \alpha_4}
    \end{gather*}
    We note that $\min(1,\eta\beta_2)>\beta_4$ and $\min(1,\eta\alpha_2)>\alpha_4$.
\end{proof}

\begingroup
\def\thetheorem{\ref{thm:flqmi_rel}}
\begin{theorem}
    Let $A$ contain at least one targeted instance ($\chi \geq 1$). Using the notations of Table~\ref{tab:variables}, the Facility Location Variant Mutual Information (FLQMI) enjoys the following bounds on $\chi$ if $\alpha_1 < \alpha_2,\beta_1 < \beta_2$:
    \begin{align*}
        \chi &\geq \frac{I_F(A;\Qcal) - \eta B\beta_1 - |\Qcal|\beta_3}{\eta(\beta_2 - \beta_1)}\\
        \chi &\leq \frac{I_F(A;\Qcal) - \eta B\alpha_1 - |\Qcal|\alpha_3}{\eta(\alpha_2 - \alpha_1)}
    \end{align*}
\end{theorem}
\addtocounter{theorem}{-1}
\endgroup
\begin{proof}
    We obtain the result by making $\alpha,\beta$ substitutions after separating the sums of FLQMI:
    \begin{gather*}
        I_F(A;\Qcal) = \sum\limits_{i\in\Qcal} \max\limits_{j\in A} s_{ij} + \eta\left(\sum\limits_{i\in A\cap\Tcal}\max\limits_{j\in\Qcal} s_{ij} + \sum\limits_{i\in A\cap\Ucal}\max\limits_{j\in\Qcal} s_{ij}\right) \leq |Q|\beta_3 + \eta\left( \chi\beta_2 + (B - \chi)\beta_1 \right)\\
        \implies\\
        \chi \geq \frac{I_F(A;\Qcal) - \eta B\beta_1 - |\Qcal|\beta_3}{\eta(\beta_2 - \beta_1)}\\
        I_F(A;\Qcal) \geq |Q|\alpha_3 + \eta \left( \chi\alpha_2 + (B - \chi)\alpha_1\right)\\
        \implies\\
        \chi \leq \frac{I_F(A;\Qcal) - \eta B\alpha_1 - |\Qcal|\alpha_3}{\eta(\alpha_2 - \alpha_1)}
    \end{gather*}
\end{proof}

\begingroup
\def\thetheorem{\ref{thm:gcmi_rel}}
\begin{theorem}
    Using the notations of Table~\ref{tab:variables}, the Graph Cut Mutual Information (GCMI) enjoys the following bounds on $\chi$ if $\gamma_1 < \gamma_2,\Delta_1 < \Delta_2$:
    \begin{align*}
        \chi &\geq \frac{\frac{1}{2\lambda |\Qcal|} I_F(A;\Qcal)-B\Delta_1}{\Delta_2 - \Delta_1}\\
        \chi &\leq \frac{\frac{1}{2\lambda |\Qcal|} I_F(A;\Qcal)-B\gamma_1}{\gamma_2 - \gamma_1}
    \end{align*}
\end{theorem}
\addtocounter{theorem}{-1}
\endgroup
\begin{proof}
    We obtain the result by making $\gamma,\Delta$ substitutions after separating the sums of GCMI:
    \begin{gather*}
        I_F(A;\Qcal) = 2\lambda\sum\limits_{i\in\Qcal}\left(\sum\limits_{j\in A\cap\Tcal} s_{ij} + \sum\limits_{j\in A\cap\Ucal} s_{ij} \right) \leq 2\lambda|\Qcal|\left( \chi\Delta_2 + (B-\chi)\Delta_1\right)\\
        \implies\\
        \chi \geq \frac{\frac{1}{2\lambda |\Qcal|} I_F(A;\Qcal)-B\Delta_1}{\Delta_2 - \Delta_1}\\
        I_F(A;\Qcal) \geq 2\lambda|\Qcal|\left( \chi\gamma_2 + (B-\chi)\gamma_1\right)\\
        \implies\\
        \chi \leq \frac{\frac{1}{2\lambda |\Qcal|} I_F(A;\Qcal)-B\gamma_1}{\gamma_2 - \gamma_1}
    \end{gather*}
\end{proof}

\begingroup
\def\thetheorem{\ref{thm:com_rel}}
\begin{theorem}
    There exist strictly monotonic increasing functions $f_l,f_h$ such that $f_l(\chi) \leq I_F(A;\Qcal) \leq f_h(\chi)$ for Concave over Modular if $\gamma_2 > \gamma_1,\gamma_4>\gamma_3$ and $\Delta_2 > \Delta_1,\Delta_4>\Delta_3$.
\end{theorem}
\addtocounter{theorem}{-1}
\endgroup
\begin{proof}
    We obtain the result by making $\gamma,\Delta$ substitutions after separating the sums of COM:
    \begin{gather*}
        I_F(A;\Qcal) = \eta\sum\limits_{i\in A\cap\Tcal} \psi\left(\sum\limits_{j\in\Qcal} s_{ij} \right) + \eta\sum\limits_{i\in A\cap\Ucal}\psi\left(\sum\limits_{j\in\Qcal} s_{ij} \right) + \sum\limits_{j\in\Qcal} \psi\left(\sum_{j\in A\cap\Tcal} s_{ij} + \sum_{j\in A\cap\Ucal} s_{ij}\right)\\
        \leq \eta\chi\psi(|\Qcal|\Delta_2) + \eta(B-\chi)\psi(|\Qcal|\Delta_1) + |\Qcal|\psi(\chi\Delta_4 + (B-\chi)\Delta_3)\\
        = \eta\chi (\psi(|Q|\Delta_2)-\psi(|Q|\Delta_1)) + \eta B\psi(|Q|\Delta_1) + |Q|\psi(B\Delta_3 + \chi(\Delta_4 - \Delta_3)) = f_h\\
        I_F(A;\Qcal) \geq \eta\chi\psi(|\Qcal|\gamma_2) + \eta(B-\chi)\psi(|\Qcal|\gamma_1) + |\Qcal|\psi(\chi\gamma_4 + (B-\chi)\gamma_3)\\
        = \eta\chi (\psi(|Q|\gamma_2)-\psi(|Q|\gamma_1)) + \eta B\psi(|Q|\gamma_1) + |Q|\psi(B\gamma_3 + \chi(\gamma_4 - \gamma_3)) = f_l
    \end{gather*}
    Per our assumptions on the $\gamma,\Delta$ parameters, we note that both $f_l,f_h$ must be strictly monotonic increasing in $\chi$, concluding the proof.
\end{proof}

\section{Proofs of Query Coverage}
\label{app:coverage}

Here, we present the proofs of Theorem~\ref{thm:flvmi_cov}, Theorem~\ref{thm:flqmi_cov}, Theorem~\ref{thm:gcmi_cov}, and Theorem~\ref{thm:com_cov} as discussed in Section~\ref{sec:coverage}. For clarity, we repeat the statement of each theorem and include their proofs here. Note that each theorem refers to Table~\ref{tab:variables}.

\begingroup
\def\thetheorem{\ref{thm:flvmi_cov}}
\begin{theorem}
    Let $A$ contain at least one targeted instance and one untargeted instance ($1 \leq \chi < B$). Using the notations of Table~\ref{tab:variables}, the Facility Location Mutual Information (FLVMI) enjoys the following bounds on $\delta_{\text{avg}}^{\Tcal\setminus A}:$
    \begin{gather*}
        \delta_{\text{avg}}^{\Tcal\setminus A} \geq \frac{I_F(A;\Qcal) - |\Ucal|\min(\eta\beta_1,1) - \chi\min(\eta\beta_2,1) + \Ocal}{|\Tcal| - \chi}\\
        \delta_{\text{avg}}^{\Tcal\setminus A} \leq \frac{I_F(A;\Qcal) - (B-\chi)\min(\eta\alpha_1,1)-\chi\min(\eta\alpha_2,1) - (|\Ucal|-B-\chi)\min(\max(\Omega_\Ucal,\Omega_{\Ucal\Tcal}),\eta\alpha_1) + \Ocal}{|\Tcal| - \chi}
    \end{gather*}
\end{theorem}
\addtocounter{theorem}{-1}
\endgroup
\begin{proof}
    As before, we first separate the sums of FLVMI, but we additionally partition $\Ucal$ across $A$:
    \begin{align*}
        I_F(A;\Qcal) = &\sum_{i\in\Ucal\cap A} \min(\max\limits_{j\in A} s_{ij}, \eta\max\limits_{j\in\Qcal} s_{ij}) + \sum_{i\in\Tcal\cap A} \min(\max\limits_{j\in A} s_{ij}, \eta\max\limits_{j\in\Qcal} s_{ij})\\
        + &\sum_{i\in\Ucal\setminus A} \min(\max\limits_{j\in A} s_{ij}, \eta\max\limits_{j\in\Qcal} s_{ij}) + \sum_{i\in\Tcal\setminus A} \min(\max\limits_{j\in A} s_{ij}, \eta\max\limits_{j\in\Qcal} s_{ij})
    \end{align*}
    The sums involving the intersection with $A$ simplify:
    \begin{align*}
        I_F(A;\Qcal) &= \sum_{i\in\Ucal\cap A} \min(1, \eta\max\limits_{j\in\Qcal} s_{ij}) + \sum_{i\in\Tcal\cap A} \min(1, \eta\max\limits_{j\in\Qcal} s_{ij})\\
        &+ \sum_{i\in\Ucal\setminus A} \min(\max\limits_{j\in A} s_{ij}, \eta\max\limits_{j\in\Qcal} s_{ij}) + \sum_{i\in\Tcal\setminus A} \min(\max\limits_{j\in A} s_{ij}, \eta\max\limits_{j\in\Qcal} s_{ij})
    \end{align*}
    We additionally can introduce $\delta_{\text{avg}}^{\Tcal\setminus A}$ by noting that the last term simply truncates elements of its sum by inducing the minimum with $\eta\max\limits_{j\in\Qcal}s_{ij}$:
    \begin{align*}
        I_F(A;\Qcal) &= \sum_{i\in\Ucal\cap A} \min(1, \eta\max\limits_{j\in\Qcal} s_{ij}) + \sum_{i\in\Tcal\cap A} \min(1, \eta\max\limits_{j\in\Qcal} s_{ij}) + \sum_{i\in\Ucal\setminus A} \min(\max\limits_{j\in A} s_{ij}, \eta\max\limits_{j\in\Qcal} s_{ij})\\ 
        &+ \sum_{i\in\Tcal\setminus A} \max\limits_{j\in A} s_{ij} - \Ocal
    \end{align*}
    Now, we make substitutions per Table~\ref{tab:variables} as before:
    \begin{gather*}
        I_F(A;\Qcal) \leq (B-\chi)\min(\eta\beta_1,1)+\chi\min(\eta\beta_2,1)-\Ocal +(|\Ucal|-B+\chi)\min(\eta\beta_1,1)+(|\Tcal|-\chi)\delta_{\text{avg}}^{\Tcal\setminus A}\\
        \implies\\
        \delta_{\text{avg}}^{\Tcal\setminus A} \geq \frac{I_F(A;\Qcal) - |\Ucal|\min(\eta\beta_1,1) - \chi\min(\eta\beta_2,1) + \Ocal}{|\Tcal| - \chi}\\
        \begin{aligned} I_F(A;\Qcal) &\geq (B-\chi)\min(\eta\alpha_1,1)+\chi\min(\eta\alpha_2,1) +(|\Ucal|-B+\chi)\min(\max(\Omega_\Ucal,\Omega_{UT}),\eta\alpha_1)\\ 
        &+(|\Tcal|-\chi)\delta_{\text{avg}}^{\Tcal\setminus A}-\Ocal\end{aligned}\\
        \implies\\
        \delta_{\text{avg}}^{\Tcal\setminus A} \leq \frac{I_F(A;\Qcal) - (B-\chi)\min(\eta\alpha_1,1)-\chi\min(\eta\alpha_2,1) - (|\Ucal|-B-\chi)\min(\max(\Omega_\Ucal,\Omega_{\Ucal\Tcal}),\eta\alpha_1) + \Ocal}{|\Tcal| - \chi}
    \end{gather*}
\end{proof}

\begingroup
\def\thetheorem{\ref{thm:flqmi_cov}}
\begin{theorem}
    Let $A$ contain at least one targeted instance ($\chi \geq 1$). Using the notations of Table~\ref{tab:variables}, the Facility Location Variant Mutual Information (FLQMI) enjoys the following bounds on $\delta_{\text{avg}}^{\Qcal}$:
    \begin{align*}
        \delta_{\text{avg}}^\Qcal &\geq \frac{I_F(A;\Qcal) - \eta(\chi\beta_2 + (B-\chi)\beta_1)}{|\Qcal|}\\
        \delta_{\text{avg}}^\Qcal &\leq \frac{I_F(A;\Qcal) - \eta(\chi\alpha_2 + (B-\chi)\alpha_1)}{|\Qcal|}
    \end{align*}
\end{theorem}
\addtocounter{theorem}{-1}
\endgroup
\begin{proof}
    The proof for average $Q$-coverage for FLQMI comes from an immediate substitution of $\delta_{\text{avg}}^{\Qcal}$:
    \begin{align*}
        I_F(A;\Qcal) = \eta\sum_{i\in A} \max\limits_{j\in Q} s_{ij} + |Q|\delta_{\text{avg}}^{\Qcal}
    \end{align*}
    As before, we substitute the notations of Table~\ref{tab:variables}:
    \begin{gather*}
        I_F(A;\Qcal) \leq \eta(\chi\beta_2 + (B-\chi)\beta_1) + |Q|\delta_{\text{avg}}^{\Qcal}\\
        \implies\\
        \delta_{\text{avg}}^{\Qcal} \geq \frac{I_F(A;\Qcal) - \eta(\chi\beta_2 + (B-\chi)\beta_1)}{|\Qcal|}\\
        I_F(A;\Qcal) \geq \eta(\chi\alpha_2 + (B-\chi)\alpha_1) + |Q|\delta_{\text{avg}}^{\Qcal}\\
        \implies\\
        \delta_{\text{avg}}^\Qcal \leq \frac{I_F(A;\Qcal) - \eta(\chi\alpha_2 + (B-\chi)\alpha_1)}{|\Qcal|}
    \end{gather*}
\end{proof}

\begingroup
\def\thetheorem{\ref{thm:gcmi_cov}}
\begin{theorem}
    Using the notations of Table~\ref{tab:variables}, the Graph Cut Mutual Information (GCMI) enjoys the following bounds on $\delta_{\text{avg}}^{\Qcal}$:
    \begin{align*}
        \delta_{\text{avg}}^\Qcal &\geq \frac{1}{2\lambda|\Qcal|} I_F(A;\Qcal) - B\Delta_1 - \chi(\Delta_2 - \Delta_1)\\
        \delta_{\text{avg}}^\Qcal &\leq \frac{1}{2\lambda|\Qcal|} I_F(A;\Qcal) - (B - 1)\gamma_1 + \gamma_2 - \chi(\gamma_2 - \gamma_1)
    \end{align*}
\end{theorem}
\addtocounter{theorem}{-1}
\endgroup
\begin{proof}
    To show average $\Qcal$-coverage bounds, the sums of GCMI are separated as follows:
    \begin{align*}
        I_F(A;\Qcal) &= 2\lambda\sum_{i\in\Qcal}\sum_{j\in A\setminus\{k_i\}} s_{ij} + 2\lambda\sum_{i\in\Qcal} \max\limits_{j\in A} s_{ij}\\
        &= 2\lambda\sum_{i\in\Qcal}\sum_{j\in A\setminus\{k_i\}} s_{ij} + 2\lambda|\Qcal|\delta_{\text{avg}}^\Qcal
    \end{align*}
    where $k_i$ denotes the most similar element in $A$ for an $i\in\Qcal$. We can further deconstruct the GCMI summation by separating the sum and introducing our $\gamma,\Delta$ substitutions as we had done in Theorem~\ref{thm:gcmi_rel}:
    \begin{gather*}
        I_F(A;\Qcal) \leq 2\lambda|\Qcal|(\chi\Delta_2 + (B-\chi)\Delta_1) + 2\lambda|\Qcal|\delta_{\text{avg}}^\Qcal\\
        \implies\\
        \delta_{\text{avg}}^\Qcal \geq \frac{1}{2\lambda|\Qcal|} I_F(A;\Qcal) - B\Delta_1 - \chi(\Delta_2 - \Delta_1)\\
        I_F(A;\Qcal) \geq 2\lambda|\Qcal|\left( (\chi-1)\gamma_2 + (B-\chi-1)\gamma_1\right) + 2\lambda|\Qcal|\delta_{\text{avg}}^\Qcal\\
        \implies\\
        \delta_{\text{avg}}^\Qcal \leq \frac{1}{2\lambda|\Qcal|} I_F(A;\Qcal) - (B - 1)\gamma_1 + \gamma_2 - \chi(\gamma_2 - \gamma_1)
    \end{gather*}
\end{proof}

\begingroup
\def\thetheorem{\ref{thm:com_cov}}
\begin{theorem}
    Let $A$ contain at least one targeted instance ($\chi \geq 1$). Then, there exist strictly monotonic $f_l,f_h$ for $i\in\Qcal$ such that the following holds for fixed $\chi$:
    \begin{align*}
        \sum_{i\in\Qcal} f_l(\max_{j\in A} s_{ij}) \leq I_F(A;\Qcal) \leq \sum_{i\in\Qcal} f_h(\max_{j\in A} s_{ij})
    \end{align*}
\end{theorem}
\addtocounter{theorem}{-1}
\endgroup
\begin{proof}
    We first separate the summations within COM:
    \begin{align*}
        I_F(A;\Qcal) = \eta \left(\sum_{i\in A\cap\Tcal} \psi(\sum_{j\in\Qcal} s_{ij}) + \sum_{i\in A\cap\Ucal} \psi(\sum_{j\in\Qcal} s_{ij})\right) + \sum_{i\in\Qcal} \psi(\sum_{j\in A\cap\Tcal\setminus\{k_i\}} s_{ij} + \sum_{j\in A\cap\Ucal} s_{ij} + \max\limits_{j\in A} s_{ij})
    \end{align*}
    where $k_i$ denotes the most similar element in $A$ for an $i\in\Qcal$. Next, we can make substitutions using the notations of Table~\ref{tab:variables}:
    \begin{align*}
        I_F(A;\Qcal) &\leq \eta\left(\chi\psi(|\Qcal|\Delta_2) + (B-\chi)\psi(|\Qcal|\Delta_1)\right) + \sum_{i\in\Qcal} \psi((\chi-1)\Delta_4 + (B-\chi)\Delta_3 + \max\limits_{j\in A} s_{ij})\\
        I_F(A;\Qcal) &\geq \eta\left(\chi\psi(|\Qcal|\gamma_2) + (B-\chi)\psi(|\Qcal|\gamma_1)\right) + \sum_{i\in\Qcal} \psi((\chi-1)\gamma_4 + (B-\chi)\gamma_3 + \max\limits_{j\in A} s_{ij})
    \end{align*}
    We can then define $f_l$ and $f_h$ as follows:
    \begin{align*}
        f_l(x) &= \frac{\eta}{|\Qcal|}\left(\chi\psi(|\Qcal|\gamma_2) + (B-\chi)\psi(|\Qcal|\gamma_1)\right) + \psi((\chi-1)\gamma_4 + (B-\chi)\gamma_3 + x)\\
        f_h(x) &= \frac{\eta}{|\Qcal|}\left(\chi\psi(|\Qcal|\Delta_2) + (B-\chi)\psi(|\Qcal|\Delta_1)\right) + \psi((\chi-1)\Delta_4 + (B-\chi)\Delta_3 + x)\\
    \end{align*}
    Hence, we have the following after substitution:
    \begin{align*}
        \sum_{i\in\Qcal} f_l(\max_{j\in A} s_{ij}) \leq I_F(A;\Qcal) \leq \sum_{i\in\Qcal} f_h(\max_{j\in A} s_{ij})
    \end{align*}
    Notably, $f_l$ and $f_h$ are strictly monotonic increasing in their arguments since $\psi$ is a strictly increasing concave function.
\end{proof}

\section{Additional Details: Query Relevance}
\label{app:detail_relevance}

In this section, we provide additional details to the analysis of the query relevance theorems discussed in Section~\ref{sec:relevance}.

\subsection{FLVMI}

The bounds in Theorem~\ref{thm:flqmi_rel} show that $\chi$ is dependent on $I_F(A;\Qcal)$ and max-case similarity assumptions on the data. However, we additionally note that since FLVMI takes a summation over all the data, the bounds are subsequently affected by potentially very large $|\Ucal|$ and $|\Tcal|$ factors of how dissimilar the query set is to the untargeted data ($\alpha_1,\beta_1$ terms) and how similar the query set is to the targeted data ($\alpha_2,\beta_2$ terms), respectively. As a result, the bounds imply that FLVMI can vary widely about the $I_F(A;\Qcal)$ value, which we empirically verify in Figure~\ref{fig:relevance}. Additionally, if $A$ is a good representation of $\Tcal \setminus A$, then this drives the $\alpha_4$ and $\beta_4$ terms to be close to $\eta\alpha_2$ and $\eta\beta_2$, respectively, which lends to an increase in the variability of $\chi$ about the $I_F(A;\Qcal)$ value. In conclusion, \textbf{the bounds suggest that FLVMI's objective value may not be well-correlated with query relevance}. Indeed, Figure~\ref{fig:relevance} shows that FLVMI does not exhibit strong correlation and that the bounds match the empirical observation. Further, Table~\ref{tab:relevance_r} shows that FLVMI has the least Spearman rank correlation across both datasets ($0.90790$ for the one-target dataset and $0.86350$ for the two-target dataset).

\subsection{FLQMI}

Again, we see from Theorem~\ref{thm:flqmi_rel} that $\chi$ depends on $I_F(A;\Qcal)$. Further, as FLQMI only pays attention to cross similarities between $A$ and $\Qcal$, the final bounds improve to only $B$ and $|\Qcal|$ factors of how dissimilar the query set is to the untargeted data ($\alpha_1,\beta_1$ terms) and how similar the query set is to the targeted data ($\alpha_2,\beta_2,\alpha_3,\beta_3$ terms), respectively. Additionally, if the targeted data and untargeted data are well-separated such that the query set only has high max-case similarity to the targeted data, the denominators of the bounds will be larger, encouraging more stable bounds. The lower variability of FLQMI's bounds suggests that \textbf{it is more suited for modeling query relevance than FLVMI. Hence, FLQMI is a theoretically sound choice for modeling query relevance in cases where the max-case similarity assumptions hold}. We empirically verify this in Figure~\ref{fig:relevance}, where FLQMI's objective value exhibits stronger correlation with $\chi$. Figure~\ref{fig:relevance} also highlights the reduced variance and tightness of the bounds, further strengthening the analysis. Lastly, we note that FLQMI exhibits a higher Spearman rank correlation than FLVMI on both datasets in Table~\ref{tab:relevance_r} ($0.99999$ for the one-target dataset and $0.97546$ for the two-target dataset).

\subsection{GCMI}

The bounds of Theorem~\ref{thm:gcmi_rel} again show a direct dependence on $I_F(A;\Qcal)$. As before, GCMI only pays attention to cross similarities; however, GCMI does not consider max-case similarity assumptions but instead considers average-case similarity assumptions ($\gamma,\Delta$ terms). As GCMI can be seen as a close analog to averaging the cross similarities (by taking the sum), the resulting bound tends to be even more stabilized around the $I_F$ value. Indeed, the $I_F$ value is normalized by the query set size and the $\lambda$ parameter, and this normalized value is now only attenuated by a $B$ factor of how dissimilar the query set is to the untargeted data ($\gamma_1,\Delta_1$ terms). As before, if the targeted data and untargeted data are well-separated such that the query set only has high average-case similarity to the targeted data (low $\gamma_1,\Delta_1$ and high $\gamma_2,\Delta_2$), the denominators of the bounds will be larger, encouraging more stable bounds. Compared to FLQMI, these bounds tend to be even less variable for the above reasons when these average-case similarity assumptions hold. \textbf{Hence, GCMI tends to be even more correlated with query relevance than FLQMI}. This also follows from Figure~\ref{fig:relevance}, where GCMI's objective value exhibits the strongest correlation with $\chi$ and the least variable and tightest bounds. Table~\ref{tab:relevance_r} also corroborates this, showing a perfect Spearman rank correlation the one-target dataset and a Spearman rank correlation of $0.97730$ for the two-target dataset.

\subsection{COM}

Unlike the previous theorems, this one establishes $\chi$'s dependence on $I_F$ by establishing a feasible range for $I_F(A;\Qcal)$ given a value of $\chi$. Importantly, this range is determined by strictly monotonic increasing functions $f_l,f_h$ of $\chi$. This property helps establish $I_F(A;\Qcal)$'s correlation with $\chi$: If $I_F(A;\Qcal)$ is made larger than the upper range determined by $f_h(\chi)$, then it must be that an increase in $\chi$ must happen due to the strictly increasing monotonicity of $f_h$. Likewise, a decrease in $\chi$ must happen if $I_F(A;\Qcal)$ is made smaller than the lower range determined by $f_l(\chi)$ due to the strictly increasing monotonicity of $f_l$. This property allows us to derive relatively tight $\chi$ bounds as shown in Figure~\ref{fig:relevance}. Hence, the statement of Theorem~\ref{thm:com_rel} establishes a dependence of COM's objective value with query relevance. The strength of this dependence is consequently determined by how far separated $f_l(\chi)$ and $f_h(\chi)$ are, which is determined by average-case similarity assumptions akin to that used in Theorem~\ref{thm:gcmi_rel}. Indeed, the upper bounding function is determined by differences in the average-case similarity upper bounds (all $\Delta$ terms), and the lower bound is determined by differences in the average-case similarity lower bounds (all $\gamma$ terms). Like GCMI, COM only pays attention to cross similarities, but it also attenuates these by a concave function $\psi$ (which is assumed to be strictly monotonic increasing). Hence, \textbf{a lot of the same behaviors that are exhibited by GCMI are also exhibited by COM to a varying degree for query relevance}. In general, our empirical analysis in Figure~\ref{fig:relevance} and Table~\ref{tab:relevance_r} shows COM's objective value to have a correlation strength roughly equivalent or better than GCMI on both datasets (perfect Spearman rank correlation on the one-target dataset and $0.99043$ Spearman rank correlation on the two-target dataset), although the bounds of Theorem~\ref{thm:com_rel} are more variable than GCMI.

\section{Additional Details: Query Coverage}
\label{app:detail_coverage}

In this section, we provide additional details to the analysis of the query coverage theorems discussed in Section~\ref{sec:relevance}.

\subsection{FLVMI}

Our first result on coverage in Theorem~\ref{thm:flvmi_cov} gives $\delta_{\text{avg}}^{\Tcal\setminus A}$, which measures average coverage on the set of targeted instances that are NOT within $A$. Our bounds show a dependence on $I_F(A;\Qcal)$. The $|U|$ factors are present within our bounds; however, we find that these bounds actually tend to be less variable than those in Theorem~\ref{thm:flvmi_rel}. This happens for a number of reasons: \textbf{1)} FLVMI's objective value depends on the whole ground set, which means that $\Tcal$ needs to be covered up to a threshold of $\Qcal$'s coverage of $\Tcal$ for $I_F(A;Q)$ to be high. \textbf{2)} If $\Ucal$ is sufficiently separated from the targeted data and $\Qcal$ represents $\Tcal$ well, then the $\alpha_1,\beta_1$ parameters (max-case similarity between $\Ucal$ and $\Qcal$) should be small, mitigating the effect of the larger $\Ucal$ terms. \textbf{3)} The denominator tends to be larger as $\chi$ is usually sufficiently less than $|\Tcal|$. \textbf{4)} The bounds include a $\Ocal$ term, which measures how much $A$ overshadows $\Qcal$ in terms of representation of $\Tcal\setminus A$ and thus makes up for variances in how well $\Qcal$ represents $\Tcal\setminus A$ and how compact $\Tcal$ is in terms of similarity. For the above reasons, \textbf{Theorem~\ref{thm:flvmi_cov} implies that FLVMI is better attuned towards query coverage}. Interestingly, when this result is combined with Theorem~\ref{thm:flvmi_rel}, FLVMI tends to exhibit extremes in behavior when considering query relevance and query coverage (or, more accurately, targeted coverage): \textbf{While FLVMI's objective value is not well correlated with query relevance, it tends to be well correlated with query coverage}. Indeed, we show that our theory mirrors empirical observation in Figure~\ref{fig:coverage} and in Table~\ref{tab:coverage_r}, where FLVMI's objective value exhibits the strongest correlation with query coverage ($0.99436$ Spearman rank correlation for the one-target dataset and $0.99300$ Spearman rank correlation for the two-target dataset). Additionally, our bounds tend to be the tightest of all our considered functions (with a slight slack in the lower bound due to the higher $\beta$ terms versus $\alpha$).

\subsection{FLQMI}

Our next result on coverage in Theorem~\ref{thm:flqmi_cov} gives delta average on $\Qcal$, which more naturally arises in FLQMI due to the objective value only depending on cross-similarities between $A$ and $\Qcal$. Our bounds show a dependence on $I_F(A;\Qcal)$ bearing close resemblance to Theorem~\ref{thm:flqmi_rel}. Our bounds have smaller $B$ and $B-\chi$ terms within the numerator that attenuate $I_F$ by how well $\Qcal$ and $\Ucal$ are separated in the max case ($\alpha_1,\beta_1$ terms) and how well $\Qcal$ and $\Tcal$ are similar in the max case ($\alpha_2,\beta_2$). Despite that $B$ and $B-\chi$ are smaller multiplicative factors than $|\Ucal|$ and $|\Tcal|$ and thus tend to be less variable with respect to the data, FLQMI's modeling of query relevance (the $\sum\limits_{i\in A}\max\limits_{j\in\Qcal} s_{ij}$ term) still introduces variances in the final bound that are not fully handled by the attentuation terms. As a result, this directly affects the bounds on query coverage, which arise from the second term ($\sum\limits_{j\in\Qcal}\max\limits_{j\in A} s_{ij}$). For these reasons, Theorem~\ref{thm:flqmi_cov} implies that \textbf{FLQMI's objective value trades off between finding good query relevance and query coverage when combined with Theorem~\ref{thm:flqmi_rel}, resulting in moderate correlation between FLQMI's objective value and both query relevance and coverage}. Indeed, Figure~\ref{fig:coverage} shows a similar bound pattern for query coverage as in Figure~\ref{fig:relevance} for query relevance. Further, our bounds tend to be tight across datasets. Lastly, Table~\ref{tab:relevance_r} and Table~\ref{tab:coverage_r} confirm that FLQMI moderates between query relevance and query coverage since its Spearman rank correlation values lie in the middle of the studied SMI variants, which empirically confirms our theory.

\subsection{GCMI}

Theorem~\ref{thm:gcmi_cov} follows the previous result by providing $\delta_{\text{avg}}^{\Qcal}$ for the same reasoning concerning cross-similarities. Akin to Theorem~\ref{thm:gcmi_rel}, the bounds are dependent on a $\frac{1}{2\lambda|\Qcal|}$ scaling of $I_F(A;\Qcal)$, which is subsequently attentuated by $B$ and $\chi$ factors of average-case similarity between $\Ucal$ and $\Qcal$ ($\gamma_1,\Delta_1$) and between $\Tcal$ and $\Qcal$ ($\gamma_2,\Delta_2$). For the above reasons, \textbf{Theorem~\ref{thm:gcmi_cov} implies that GCMI can have wide variance in query coverage with respect to its objective value}. We empirically confirm this in Figure~\ref{fig:coverage}, which shows weak correlation between GCMI's objective value and $\delta_{\text{avg}}^\Qcal$. Figure~\ref{fig:coverage} additionally shows that these bounds can have very wide variance and, in some cases, are reasonably tight to the data.  Interestingly, we also see a severe degradation in Spearman rank correlation in Table~\ref{tab:coverage_r} moving from the one-target dataset to the two-target dataset ($0.87379$ to $0.64170$), which lends to poor average-case similarity guarantees and also shows GCMI's inability to effectively model query coverage. Hence, we note that GCMI behaves on the opposite end of the spectrum from FLVMI: \textbf{GCMI's objective value is highly correlated with query relevance but is weakly correlated with query coverage}.

\subsection{COM}

Theorem~\ref{thm:com_cov} follows the same reasoning as Theorem~\ref{thm:com_rel}; however, unlike both the theorems of this section and Theorem~\ref{thm:com_rel}, the mapping between $\delta_{\text{avg}}^\Qcal$ and $I_F(A;\Qcal)$ is more implicit. Usually, a high $\delta_{\text{avg}}^\Qcal$ value implies that each of the max terms contribution towards that value are also high. However, increases in $\delta_{\text{avg}}^\Qcal$ do not necessarily imply increases in $\sum\limits{i\in\Qcal} f_l (\max\limits_{j\in A} s_{ij})$ or $\sum\limits{i\in\Qcal} f_h (\max\limits_{j\in A} s_{ij})$ since some of the max terms may have \emph{decreased}, which are then attenuated by the strictly monotonic increasing concave function $\psi$. As such, the statement of Theorem~\ref{thm:com_cov} has a slightly weaker guarantee on the correlation between $I_F$ and the target metric (query coverage in this case) than in Theorem~\ref{thm:com_rel}. However, for sufficiently large increases in $\delta_{\text{avg}}^\Qcal$, it is reasonable to expect that all max terms would increase, lending to some implicit correlation between $I_F(A;\Qcal)$ and $\sum\limits{i\in\Qcal} f_l (\max\limits_{j\in A} s_{ij})$. Per Theorem~\ref{thm:com_rel}'s previous argument, the strength of the correlation is also dependent on how close $f_l(x)$ and $f_h(x)$ are. Per Theorem~\ref{thm:com_cov}'s proof (in Appendix~\ref{app:coverage}), we see that this depends on how far separated all the $\gamma$ parameters are from the $\Delta$ parameters, which reflect average-case similarities between $\Qcal$ and both $\Ucal,\Tcal$. However, such differences may be exacerbated by the concavity of $\psi$, lending to potentially more variability. For the above reasons, \textbf{Theorem~\ref{thm:com_cov} implies that COM can potentially have wide variance in query coverage with respect to its objective value}. We empirically confirm this in Figure~\ref{fig:coverage}, which shows weak correlation akin to GCMI (COM also has similar Spearman rank correlation as GCMI in Table~\ref{tab:coverage_r}). Indeed, these variances may even be exacerbated further by $\psi$. Unlike other functions, COM does not have bounds shown in Figure~\ref{fig:coverage} due to the implicit mapping discussed previously.

\end{document}